%% file: main.tex
\documentclass[conference]{IEEEtran}
\IEEEoverridecommandlockouts
\usepackage{cite}
\usepackage{amsmath,amssymb,amsfonts}
\usepackage{siunitx}
\usepackage{graphicx}
\usepackage{textcomp}
\usepackage{xcolor}
\usepackage{algorithm} 
\usepackage[margin=0.75in]{geometry}

\usepackage[hidelinks]{hyperref}
\usepackage{algpseudocode} 
\usepackage{booktabs, adjustbox}
\usepackage{multirow,enumitem}
\begin{document}

\title{\Large \bf PoseINN: Realtime Visual-based Pose Regression and Localization \\with Invertible Neural Networks\\
{}
\thanks{}
}

\author{Zirui Zang, Ahmad Amine, Rahul Mangharam\vspace{-2.5\baselineskip}
\thanks{All authors are with the University of Pennsylvania, Department of Electrical and Systems Engineering, 19104, Philadelphia, PA, USA. Emails: \{\tt\footnotesize zzang, aminea, rahulm\}@seas.upenn.edu}
  
}

\maketitle

\input{abstract}
\input{intro.tex}

\input{related.tex}
\input{method.tex}
\input{exp.tex}
\input{discussion.tex}

\bibliographystyle{IEEEtran}
\bibliography{IEEEabrv,bib/main,bib/others,bib/nerf,bib/pose}

\end{document}

%% file: abstract.tex
\begin{abstract}
Estimating ego-pose from cameras is an important problem in robotics with applications ranging from mobile robotics to augmented reality. While SOTA models are becoming increasingly accurate, they can still be unwieldy due to high computational costs. In this paper, we propose to solve the problem by using invertible neural networks (INN) to find the mapping between the latent space of images and poses for a given scene. Our model achieves similar performance to the SOTA while being faster to train and only requiring offline rendering of low-resolution synthetic data. By using normalizing flows, the proposed method also provides uncertainty estimation for the output. We also demonstrated the efficiency of this method by deploying the model on a mobile robot.
\end{abstract}

%% file: intro.tex
\section{Introduction}

Visual pose regression is the task of finding camera poses of images within a trained environment. The matured geometric-based pipeline\cite{schonberger2016structure,fisher2021colmap,sattler2012improving} can lead to expensive computation and long latency. On the other hand, learning-based pose regression\cite{kendall2015posenet,wu2017delving,moreau2022coordinet,melekhov2017image,liu2018intriguing} has improved efficiency but can be cumbersome to deploy due to their low accuracy and long training time. Recently, with aid from neural radiance fields (NeRF)\cite{mildenhall2021nerf}, learning-based pose regression methods have greatly improved their accuracy\cite{moreau2022lens,chen2022dfnet}. Direct feature-matching with online-rendered images and synthetic training data generation are two ways people use NeRF to improve pose regression. Despite that, these efforts either need online rendering with NeRF or long-time synthetic data preparation. 

\begin{figure}[t]
\centering
\includegraphics[width=1\columnwidth]{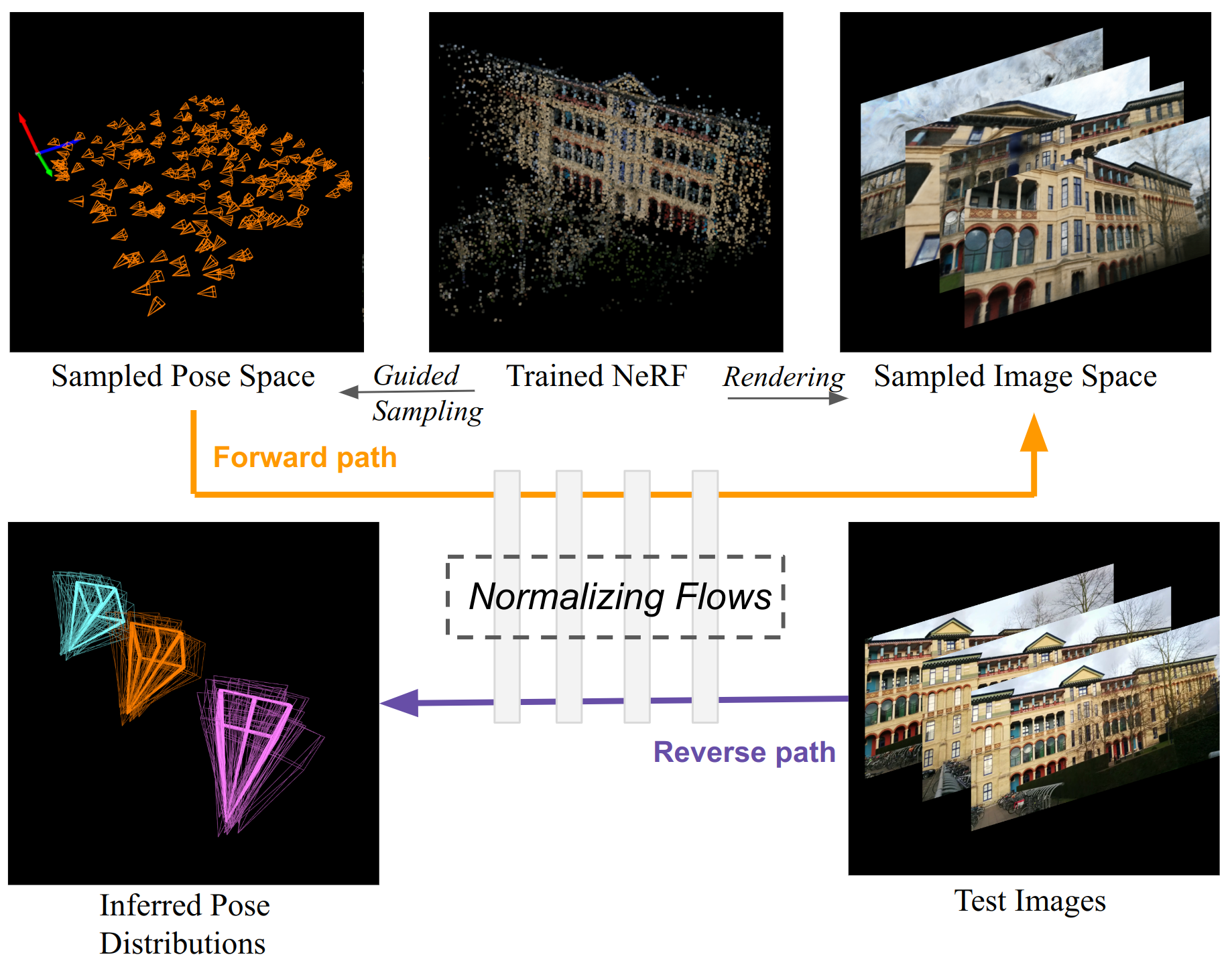}
\caption{We propose to learn a mapping between the latent space of the images and camera poses in an environment with an invertible neural network. We use NeRF to guide camera pose sampling and render synthetic images. Evaluating the reverse path of the INN outputs the full posterior distribution of camera poses given a test image.}
\label{fig_cover}
\vspace{-15pt}
\end{figure}

To address these limitations, we propose to use NeRF to render a large number of low-resolution images and view the problem as finding a mapping between the distributions of camera poses and images with normalizing flows. NeRF enabled us to conveniently sample in the image space and fully utilize the 3D spatial information embedded in the training dataset. During the evaluation, we can find the full posterior distribution of poses given the images by sampling the latent space of the INN. We summarize our contributions as the following:

\begin{enumerate}
    \item We extend Local\_INN\cite{zang2022local_inn} from LiDAR to cameras, which expands the usability for real robots. The method is tested on common benchmark datasets and the performance is on par with state-of-the-art.
    \item We realize a fast data preparation pipeline with NeRF\cite{mildenhall2021nerf,nerfstudio}, which further lowers the deployment burden.
    \item We demonstrate the balance of performance and efficiency of the proposed method by deploying it on a real mobile robot.
\end{enumerate}

%% file: related.tex
\section{Related Work}

\subsection{Visual Pose Regression}

The pioneering work in pose regression by PoseNet\cite{kendall2015posenet} used simple CNN + average pooling layers to regress camera poses. Since that, the state-of-the-art (SOTA) was yearly refreshed by people trying more complex neural network architectures, such as using separate outputs for position and orientation\cite{wu2017delving}, translational invariant layers\cite{liu2018intriguing,moreau2022coordinet}, or LSTM\cite{wang2020atloc}, auto-encoders\cite{shavit2022camera}, transformers\cite{shavit2021learning}, etc. To show the effectiveness of our method, we are using an encoder that is also simply CNN + average pooling and yet still performs on par with the SOTAs.

Recently, pose regression tasks benefited from NeRF's ability to render photo-realistic images from novel camera poses. LENS\cite{moreau2022lens} augments the training data by rendering synthetic images with a trained NeRF-W from a grid-based novel pose sampling. The limitation of LENS is the days-long training time and high-resolution image rendering time. On the other hand, Direct-PoseNet\cite{chen2021direct} uses a photometric loss to compare the test images with NeRF-rendered images at test poses. DFNet\cite{chen2022dfnet} improved that with direct feature-matching in the feature space instead of pixel-value space. However, these methods require expensive online rendering from NeRF. 
Different from the SOTA's complex approach, we claim that offline rendering of many low-resolution images is enough to perform the localization. Given a test image, our method also produces the posterior distribution of camera poses, which can be used as uncertainty estimations \cite{kendall2016modelling,moreau2022coordinet,zangeneh2023probabilistic} to improve robustness and deployability.

\subsection{Normalizing Flows}

Normalizing flows use a series of bijective transformations to map a source distribution to a target distribution. They provide efficient density estimation\cite{dinh2016density,kingma2018glow} and sampling of the target distribution. Ardizonne et al.\cite{ardizzone2018analyzing,ardizzone2019guided,winkler2019learning} proposed a framework for using normalizing flows to solve ambiguous inverse problems. The use of INNs in solving inverse problems has been applied to various fields\cite{adler2019uncertainty,ardizzone2018analyzing,WehRud2021,zhao2021invertible}. Recently, Local\_INN\cite{zang2022local_inn} has shown the effectiveness of INNs in performing robot localization, which is naturally an ambiguous inverse problem. However, \cite{zang2022local_inn} uses LiDAR ranges, which can be simulated with high fidelity given an occupancy map of the environment. Although LiDAR data provide reliable distance measurements, the sensor is expensive and lacks color information about the world. We extend that framework for visual 6DoF pose regression which is a more common problem. For that we developed a synthetic pose sampling policy with NeRF guidance.

%% file: method.tex
\section{Method}
\begin{figure}[h]
\centering
\includegraphics[width=1\columnwidth]{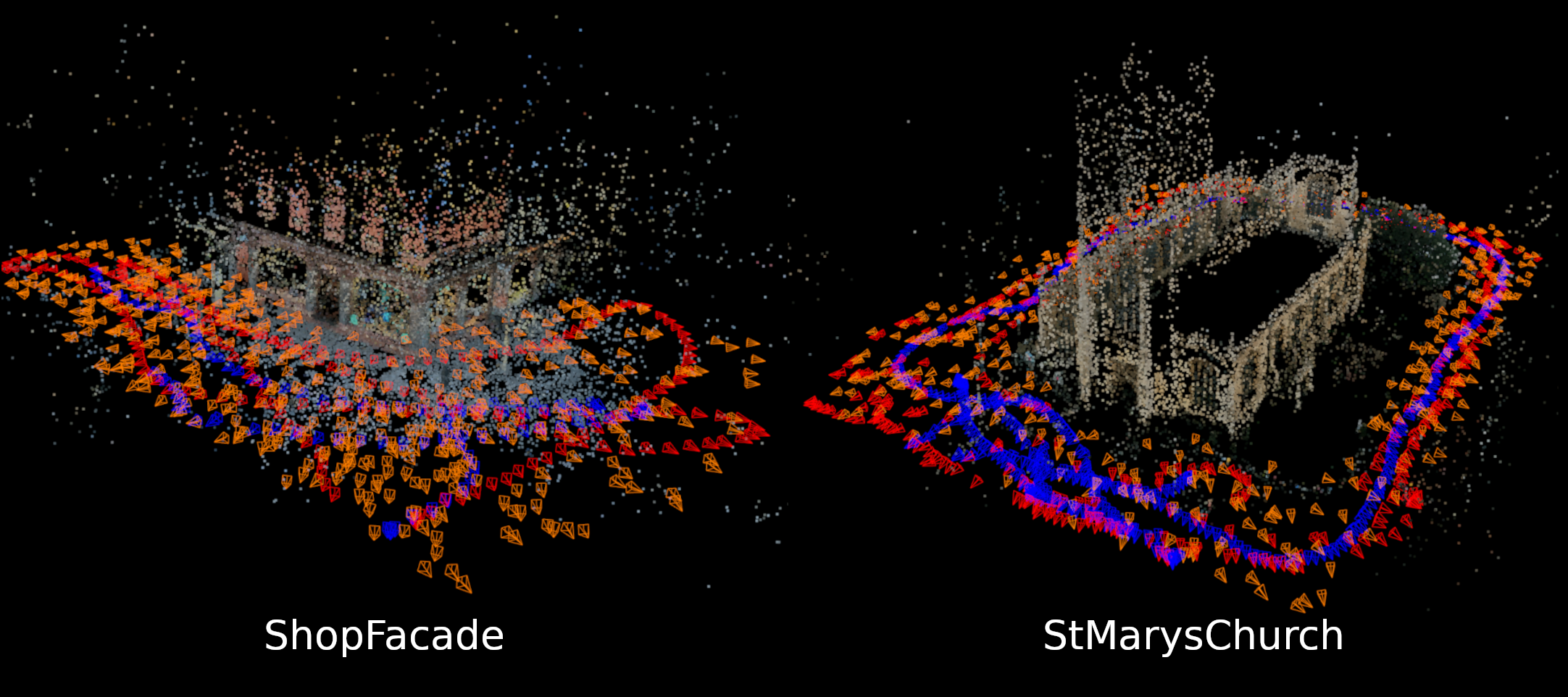}
\caption{Sampling of Novel Camera Poses. Point clouds represent high-density points in the environment. Small pyramids represent \textcolor{red}{training poses}, \textcolor{blue}{testing poses}, and \textcolor{orange}{sampled poses}.}
\label{sampling}
\vspace{-10pt}
\end{figure}

Our approach to visual pose regression is to view it as finding a mapping from the distribution of the image to that of camera poses. Effectively sampling enough corresponding data points in both distributions is the key to finding such mapping. We train a Neural Radiance Fields (NeRF) model of the environment and use it to render images at randomly sampled novel camera poses as in Fig. \ref{sampling}. We propose a random camera pose sampling and synthetic image rendering pipeline that is fundamental to the final pose regression result.

Once we have generated enough image samples, based on \cite{zang2022local_inn}, we use normalizing flows combined with variational autoencoder (VAE) to learn the mapping from pose to images. To reduce the dimensionality of image data, we use a VAE to encode images into a latent space. Then, we use coupling-based normalizing flows to learn the mapping from encoded images to poses. The latent space of the normalizing flows is sampled according to a normal distribution during training. During the evaluation, we evaluate only the encoder of the VAE and the reverse path of the normalizing flows with repeatedly sampled INN latent space to reveal the full posterior distribution of the poses given an input image.

\subsection{Generate Synthetic Views with NeRF}

A NeRF model stores 3D spatial information of an environment implicitly within two neural networks: A density MLP and a color MLP, which can be queried for any point in the continuous 3D space. Images can be rendered by tracing rays from the environment to the image plane, integrating the density and color information provided by the two MLPs. We train a NeRF model with a set of images with known camera poses, optimizing the rendering loss. However, if the learned density and color information is noisy or missing, the rendered images will contain artifacts or be a complete mess. Therefore, selecting suitable rendering poses while sufficiently sampling the wanted 3D space is challenging.

We used nerfacto\cite{nerfstudio} as our NeRF model. After training the model, we output a sparse point cloud from NeRF by thresholding the density of the environment. To generate novel camera poses, we first uniformly randomly sample positions in the region. The orientations of these sampled camera poses are given as $R_\text{noise} R_\text{training}^\text{rand}$, where $R_\text{training}^\text{rand}$ is a randomly picked camera orientation from the training set and $R_\text{noise}$ is an added perturbation. We generated random rotation $R_\text{noise}$ for up to 3.6 degrees using \cite{arvo1992fast}.

For each sampled camera pose, we verify that we have sufficient spatial information by finding a subset $\mathcal{P}_{\text{in-view}}$ of the NeRF point cloud that is within the field of view (in-view) of the sample camera. We want every sampled camera pose to have enough $\mathcal{P}_{\text{in-view}}$, i.e. enough density information for rendering, and not blocked by a very close point in $\mathcal{P}_{\text{in-view}}$. We then filter out the sampled camera poses according to the following three rules:

\begin{itemize}
    \item The distance $\delta_{\text{training}}$ from the sampled pose to the nearest pose in the training set cannot be larger than 0.5 meters.
    \item For $N_{\text{in-view}} = |\mathcal{P}_{\text{in-view}}|$, we first find the range of $N_{\text{in-view}}$ of the poses in training set. Then we limit the $N_{\text{in-view}}$ of sampled poses according to that range.
    \item The distance $\delta_{\text{in-view}}$ from the sampled pose to the nearest point in $\mathcal{P}_{\text{in-view}}$ is also limited with the range of $\delta_{\text{in-view}}$ of the poses in training set.
    
\end{itemize}
Because we use a sparse point cloud, we can sample 50k poses within minutes. Synthetic images at the sampled poses are then rendered with the trained NeRF model.

\subsection{Learning the Pose-Image Mapping}
\begin{figure*}[h]
\centering
\includegraphics[width=1.6\columnwidth]{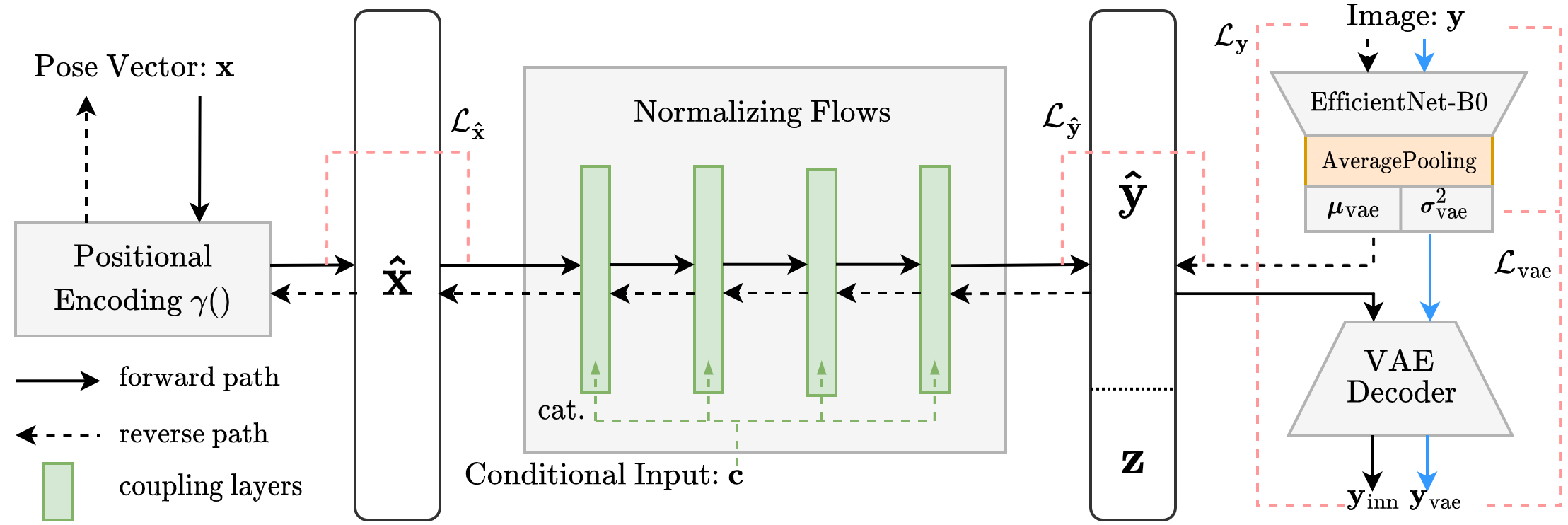}
\caption{Network Structure of the PoseINN. The forward path (solid) is from pose to image. The reverse path (dashed) is from image to pose.}
\label{fig_local_inn}
\vspace{-10pt}
\end{figure*}

Normalizing flows are a series of transformations that are mathematically invertible and with learnable parameters. Fig. \ref{fig_local_inn}, shows the structure of the network. The normalizing flows side of the network is identical to \cite{zang2022local_inn}, please refer to that paper for details. We use Real-NVP\cite{dinh2016density,kingma2018glow} for its efficiency, which uses affine coupling blocks to achieve invertibility.
$\mathbf{c}$ is the optional conditional input\cite{ardizzone2019guided}. For a fair comparison with other methods, we don't use $\mathbf{c}$ for the absolute pose regression experiments. It's only for real robot localization experiments.  Normalizing flows require the input and output to have the same dimension due to their invertibility. The 6DoF camera poses, $\mathbf{x} = [x, y, z, \theta_z, \theta_x, \theta_y]$ are augmented with Positional Encoding \cite{vaswani2017attention} \cite{mildenhall2021nerf} from $\mathbb{R}^6$ to $\mathbb{R}^{12L}$.
\begin{equation}
\begin{split}
\gamma(p)= (&\sin(2^0\pi p), \cos(2^0\pi p), \dotsc,\\  &\sin(2^{L-1}\pi p), \cos(2^{L-1}\pi p)).
\end{split}
\end{equation}
We use $L=5$ for camera poses and the output is concatenated with the original 6-dimensional pose to form an input $\mathbf{\hat{x}} \in \mathbb{R}^{12L+6}$ for the INN. On the image side, we use a VAE to encode the image $\mathbf{y}$ into $\mathbf{\hat{y}} \in \mathbb{R}^{12L}$, which is concatenated with a 6-dimensional latent vector $\mathbf{z} \sim \mathcal{N}(0,\,1)$ to form the output of the INN. Different from \cite{zang2022local_inn}, in the VAE encoder, we use a pre-trained EfficientNet-B0 backbone\cite{tan2019efficientnet} connected with an average-pooling layer to output one number for each feature channel. At test time, we can sample the latent vector to reveal the full posterior distribution of the pose given an image \cite{ardizzone2019guided}.


We train the network the same way as in \cite{zang2022local_inn}, where with each batch of data, we evaluate both the forward and reverse paths of the network and losses are added together before an optimizer step. To handle the 6DoF poses more efficiently, we used the geodesic distance\cite{zhou2019continuity} $\mathcal{L}_{\text{geo}}$ between two rotations:
\begin{equation}
\mathcal{L}_{\text{geo}} = \text{cos}^{-1}((tr(M_{\text{pred}} M_{\text{gt}}^{-1}) - 1)/2).
\end{equation} The EfficientNet backbone in the VAE is loaded with pre-trained weights when initialized and also optimized with the rest of the network in training. 

%% file: exp.tex
\section{Experiments}
We validated our method with two types of tasks. To directly compare it with other pose regression methods, we tested on public absolute pose regression datasets. We also deployed a sequential version on a mobile robot to show the performance of our method on an embedded platform.

\subsection{Camera Pose Regression on Public Dataset}

\begin{table}[h]
\centering
\vspace{-10pt}
\caption{Data Generation Strategy Comparison \\(Error Data from 7Scene)}
\begin{tabular}{ l | c c c c }
\toprule 
\shortstack{Model \\ (backbone)}
& \shortstack{Pose Error \\ (m/$^\circ$)} 
& \shortstack{Synthetic \\ Resolution} 
& \shortstack{Rendering \\ Cost} 
& \shortstack{Generation \\ Mode}\\
\midrule 
LENS(EB3)
& $0.08/3.00$ 
& High 
& Expensive 
& Offline\\

DFNet(EB0)
& $\mathbf{0.08}/3.47$  
& Low
& Cheap 
& Online \\

Ours(EB0)
& $0.09/\mathbf{2.65}$ 
& \textbf{Low} 
& \textbf{Cheap}
& \textbf{Offline} \\

\bottomrule
\end{tabular}
\label{table_strategy}
\vspace{-10pt}
\end{table}

With the 7scene\cite{glocker2013real} dataset, we trained the nerfacto model for 50k epochs, which takes about 20 mins on our setup with an NVIDIA A6000 GPU. Then 50k synthetic camera-pose images are rendered for each scene, which takes about 40 mins. The rendering resolution for the 7scene dataset is 160x120. The original training set images are then mixed with the rendered images and resized to 128x128 for training the INN. We trained the network for 300 epochs with batch size 200 and a learning rate of 5e-4 exponentially decaying to 5e-5, which takes around 8 hours. Table \ref{table_strategy} shows a comparison of the data generation strategy with LENS\cite{moreau2022lens} and DFNet\cite{chen2022dfnet}. The inputs for the other two methods are from \cite{chen2022dfnet}. Our strategy is the most efficient while outputting on-par results.



\subsection{Visual Localization on Real-world Mobile Robot}

\begin{table*}[h]
\centering
\vspace{-10pt}
\setlength{\tabcolsep}{5pt}

\caption{Median Localization Errors with 2D LiDAR vs. Camera}

\begin{tabular}{c l c c c c}
\toprule 
\multicolumn{1}{c}{Experiment Platform} &
\multicolumn{1}{c}{} & 
\multicolumn{2}{c}{Indoor} & 
\multicolumn{2}{c}{Outdoor}  \\ 

\midrule
\multicolumn{1}{c}{\adjincludegraphics[valign=c,width=0.54\columnwidth]{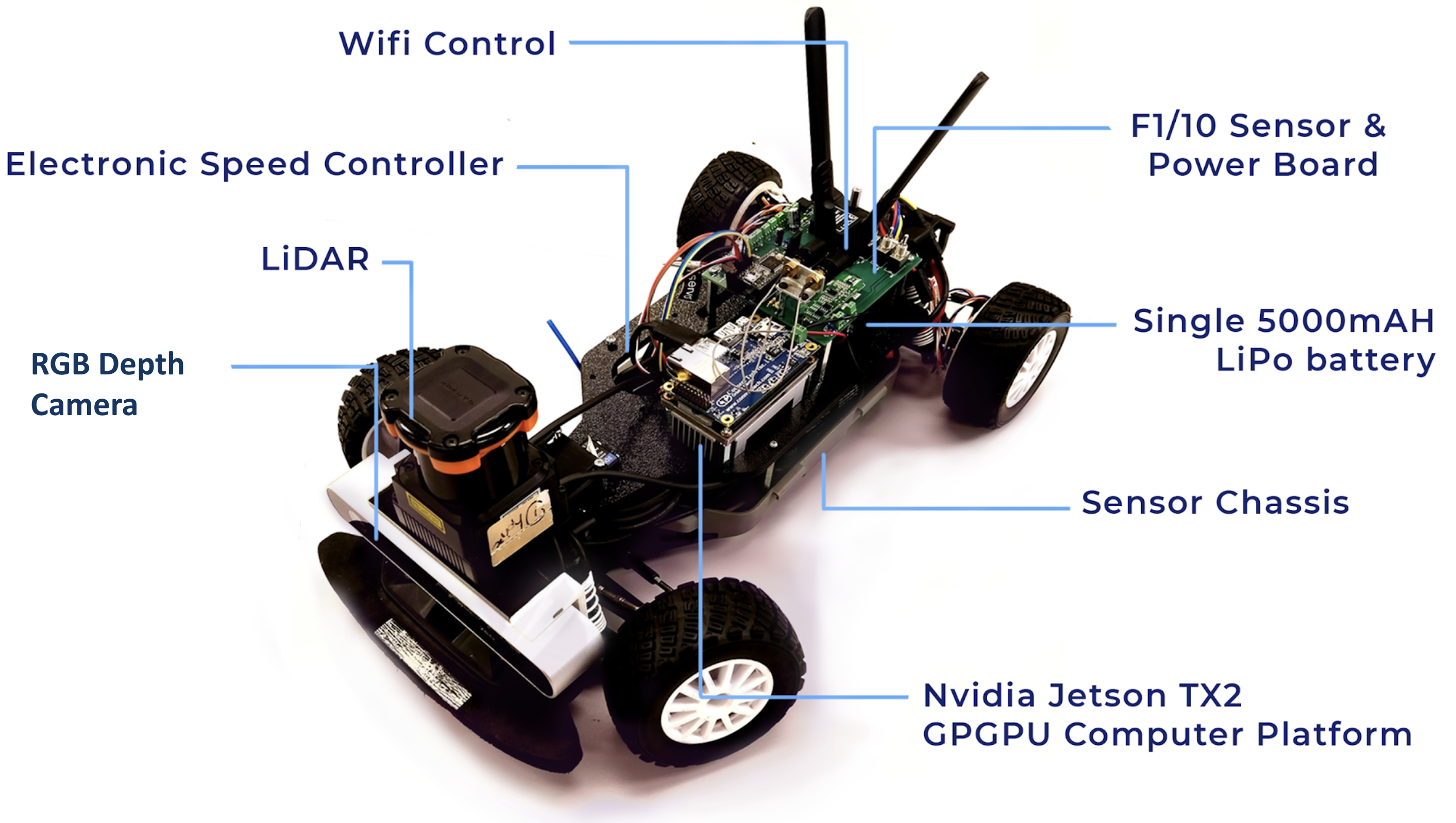}} &
\shortstack{\textcolor{blue}{train trajectory}\\
\textcolor{red}{test trajectory}\\
\textcolor{orange}{sampled points}}
& \multicolumn{2}{c}{\adjincludegraphics[valign=c,width=0.55\columnwidth]{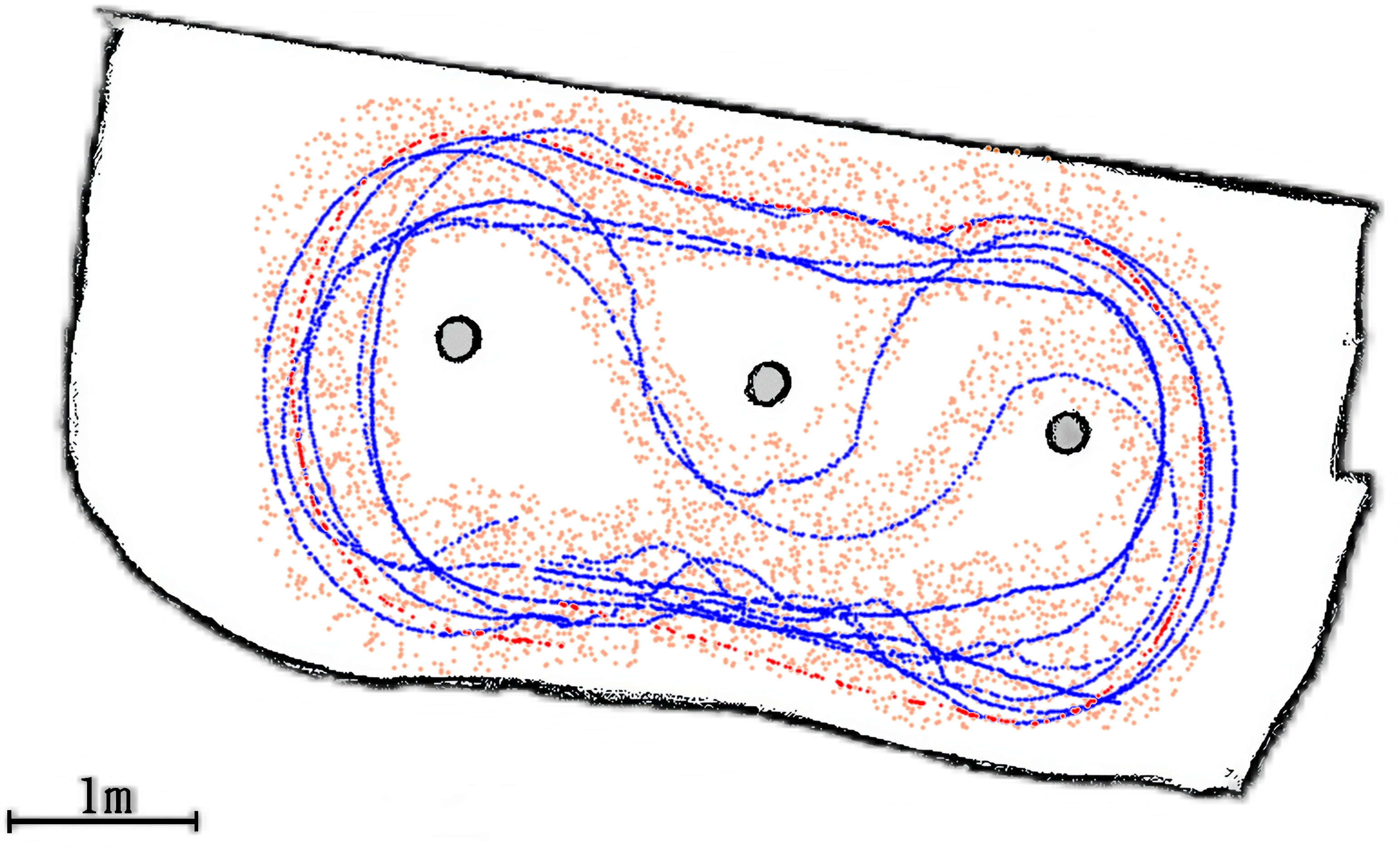} }
& \multicolumn{2}{c}{\adjincludegraphics[valign=c,width=0.55\columnwidth]{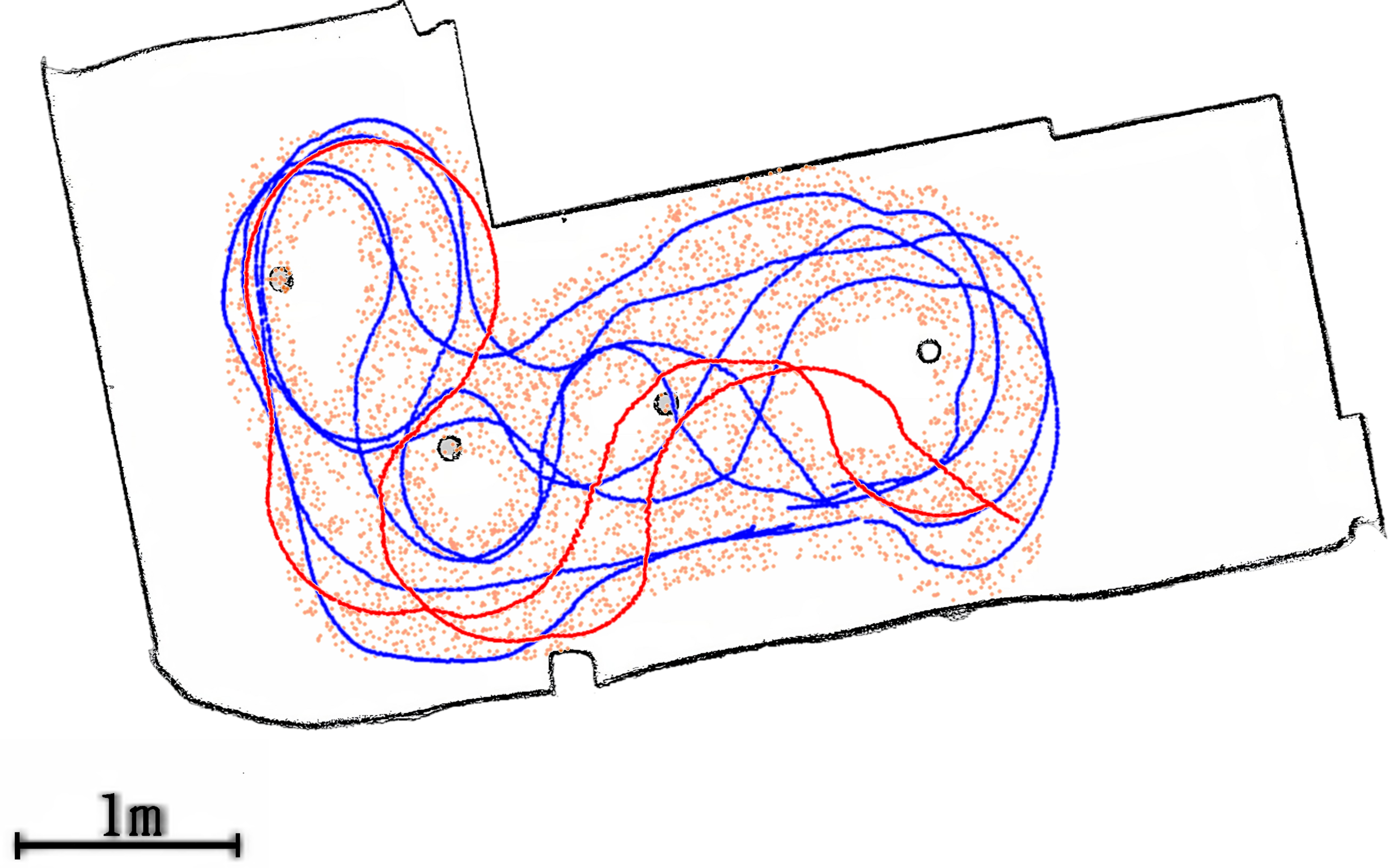} }
\\ 

&
& \multicolumn{2}{c}{($xy$[m], $\theta$[$^\circ$])}
& \multicolumn{2}{c}{($xy$[m], $\theta$[$^\circ$])} 
\\ \midrule
&
Online PF (45Hz)
& \multicolumn{2}{c}{$\mathbf{0.01}, 0.23$} 
& \multicolumn{2}{c}{$\mathbf{0.02},\mathbf{0.36}$}   \\

&
PoseINN (\textbf{154Hz})
& \multicolumn{2}{c}{$0.02,0.31$}
& \multicolumn{2}{c}{$0.12,0.72$}  \\

&
PoseINN + EKF
& \multicolumn{2}{c}{$0.02,\mathbf{0.22}$}
& \multicolumn{2}{c}{$0.10,0.65$} 

\\ \bottomrule
\end{tabular}
\label{table_2d}
\vspace{-15pt}
\end{table*}


\begin{figure}[h]
    \centering
    \includegraphics[width=0.9\columnwidth]{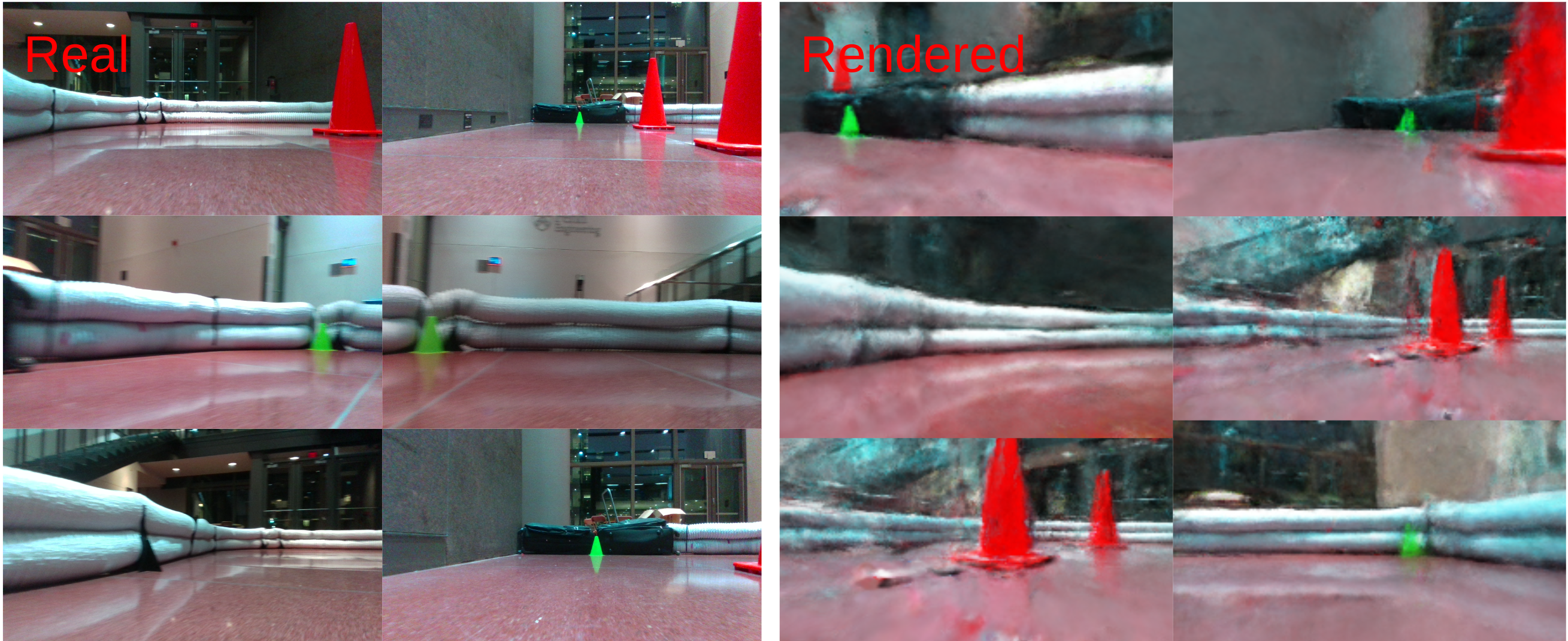}
    \includegraphics[width=0.9\columnwidth]{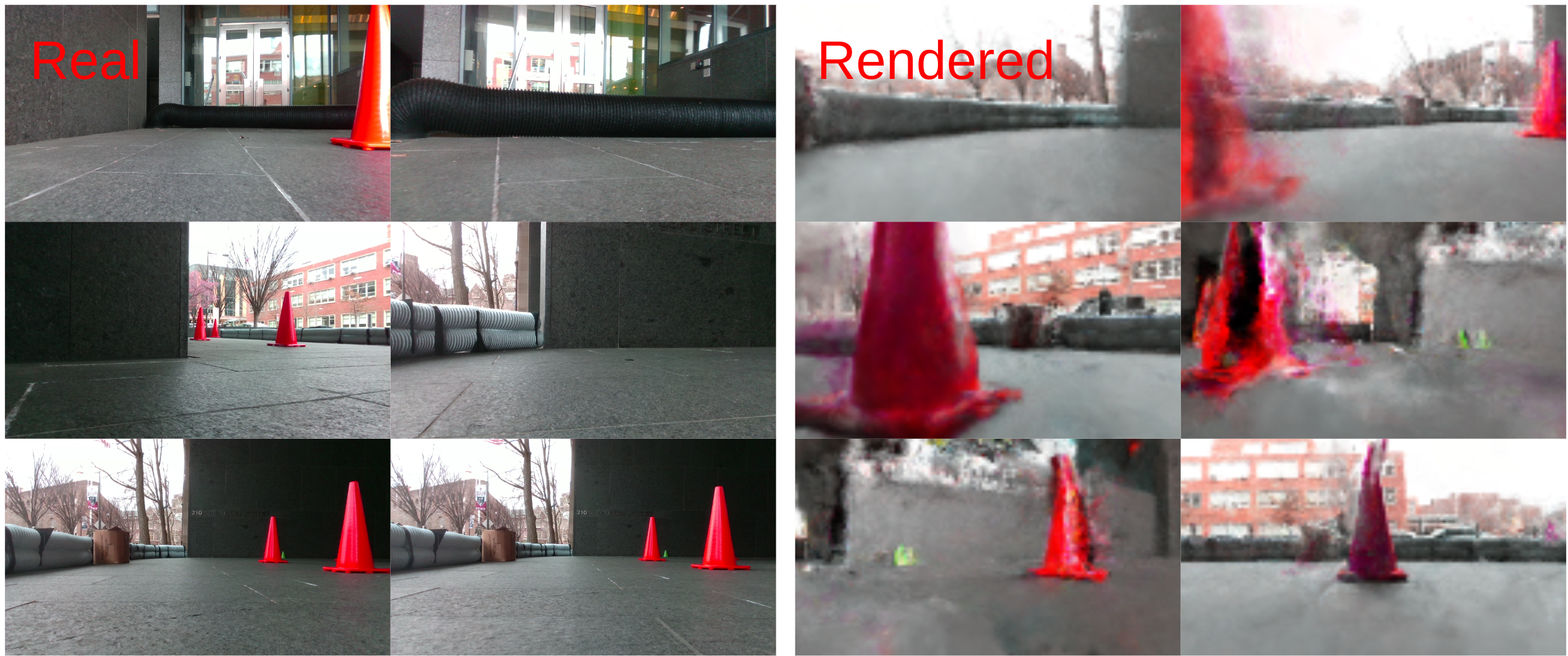}
    \caption{Examples of training and rendered images in real-world testing (Up: Indoor, Down: Outdoor)}
    \label{fig:exp_renders_out}
    \vspace{-5pt}
\end{figure}

With a small network size, PoseINN is suitable for embedded platforms. To demonstrate that, we deployed PoseINN on an \href{https://f1tenth.org/}{F1TENTH} racecar\cite{pmlr-v123-o-kelly20a}, which is a 1/10 scale autonomous racing car equipped with a Hokuyo 30LX LiDAR, an RGB camera, and an NVIDIA Jetson Xavier NX. We used LiDAR to collect ground truth poses for training images and used the camera for localization tests. For the 2D localization experiment, we train for 3 degrees of freedom: xy positions and the car's heading. Similar to \cite{zang2022local_inn}, the network architecture we used for the 2D localization experiments takes a rounded previous state of the mobile robot as conditional input $\mathbf{c}$, which is encoded by a separate MLP. This one-step historical information makes the inverse problem easier and it's used in traditional robot localization methods like particle filters\cite{dellaert1999monte, thrun2002probabilistic}.

We set up an indoor and an outdoor experiment. The maps shown in the Table \ref{table_2d} are captured with LiDAR scan using ROS SLAM toolbox. We use an offline particle filter\cite{walsh17} with an infinite computation budget for ground truth poses and training data for NeRF. An online version of the particle filter with fewer particles is used as the baseline comparison. Training and testing trajectories are also shown on the map. We capture RGB images as the car navigates along the trajectories. Fig. \ref{fig:exp_renders_out} shows the training and rendered images. We can see even without the super-accurate image renderings, the trained model is still able to provide localization. 

The translation and rotation error results in Table \ref{table_2d} show that when the training data sufficiently cover the test trajectory, this method can provide localization comparable to LiDAR-based PF. When the test trajectory moves outside the sampled zone, then the performance drops. As for runtime on the Jetson Xavier NX, PoseINN runs at 154Hz while evaluating batches with 50 randomly sampled $\mathbf{z}$ for uncertainty estimation, whereas the compared online particle filter runs at 45Hz.




\subsection{Uncertainty Estimation}

\begin{table}[h]
\centering
\vspace{-10pt}
\caption{Output Filtering with Uncertainty Estimation \\(Error Data from Cambridge\cite{kendall2015posenet})}
\begin{tabular}{ l | c c }
\toprule 
($xy$[m], $\theta$[$^\circ$]) & Raw Mean Error & With Filtering\\
\midrule 
Kings
& $0.93, 1.02$ 
& $\textbf{0.58}, \textbf{0.96}$  \\

Hospital
& $0.87, 1.14$ 
& $\textbf{0.64}, \textbf{0.97}$ \\

Shop
& $0.59, 5.00$ 
& $\textbf{0.20}, \textbf{1.04}$ \\

Church
& $0.81, 2.43$ 
& $\textbf{0.52}, \textbf{1.21}$ \\
\midrule 

Average
& $0.84, 2.55$ 
& $\textbf{0.68}, \textbf{1.87}$ \\
\bottomrule
\end{tabular}
\label{table_uncertainty}
\end{table}

We can then calculate the variance of the output distribution as uncertainty estimations. To demonstrate the effectiveness, we use the covariance of the 2D localization results with an Extended Kalman Filter (EKF) to fuse the output with odometry data from the mobile robot, which improves the accuracy. For the 3D pose regression experiments, we show that filtering the inferred poses with their variance reduces noise levels. In Table \ref{table_uncertainty}, we show the \textit{average error} of the raw outputs of PoseINN on the left. We then filter out outputs with variance values larger than the median variance value of the testing set. The average errors of outputs after filtering are in the right column. Because average values can be influenced by extreme values, a large improvement shows the output is more robust.

%% file: discussion.tex
\section{Discussion \& Limitations}

Using NeRF to efficiently sample camera poses and RGB images in an environment, we reduce the problem of pose regression into learning a mapping between two distributions. Results in Table \ref{table_strategy} show that with a large amount of lower-resolution rendering, we can achieve the same performance without using more complex methods or higher resolutions as in the compared methods. Results in Table \ref{table_2d} show the proposed method is very efficient and can provide accurate localization if proper training data is provided. The uncertainty estimation that naturally comes with the normalizing flows also makes it suitable for deployment on robot platforms.

Some limitations remain in this work. First, we didn't deal with the domain gap between NeRF-rendered images and the real images that change dramatically with the weather, camera parameters, etc. We tried to have the VAE reconstruct rendered images from real images, but the effect was not prominent. Second, we can see from our experiment that better-covered training data is crucial for the final results. Although our camera pose sampling pipeline, reduced instances of bad renderings, having a more deeply related rendering pipeline will be very helpful.

\section{Conclusion}
We showcase how this\cite{zang2022local_inn} invertible neural network architecture can be used for image-based localization at SOTA performance by only changing an image encoder. To achieve that, we used NeRF as a camera simulator to efficiently sample images within an environment. The efficiency and robustness of the model are illustrated by deploying it on an embedded mobile robot.